\documentclass[10pt,twocolumn,letterpaper]{article}

\usepackage{cvpr}
\usepackage{times}
\usepackage{epsfig}
\usepackage{graphicx}
\usepackage{amsmath}
\usepackage{amssymb}

\usepackage{booktabs}

\usepackage[breaklinks=true,letterpaper=true,colorlinks,bookmarks=false]{hyperref}

\DeclareMathOperator{\sgn}{sgn}
\DeclareMathOperator{\norm}{norm}
\DeclareMathOperator{\iou}{IoU}

\cvprfinalcopy % *** Uncomment this line for the final submission

\def\cvprPaperID{13} % *** Enter the CVPR Paper ID here

\ifcvprfinal\fi
\begin{document}

%%%%%%%%% TITLE
\title{Decoupling Video and Human Motion: Towards Practical Event Detection in Athlete Recordings}

\author{Moritz Einfalt and Rainer Lienhart\\
University of Augsburg\\
Augsburg, Germany\\
{\tt\small \{moritz.einfalt, rainer.lienhart\}@informatik.uni-augsburg.de}
% \and
% Rainer Lienhart\\
% University of Augsburg, Germany\\
%{\tt\small rainer.lienhart@informatik.uni-augsburg.de}
}

\maketitle
\thispagestyle{empty}

%%%%%%%%% ABSTRACT
\begin{abstract}
 In this paper we address the problem of motion event detection in athlete recordings from individual sports. In contrast to recent end-to-end approaches, we propose to use 2D human pose sequences as an intermediate representation that decouples human motion from the raw video information. Combined with domain-adapted athlete tracking, we describe two approaches to event detection on pose sequences and evaluate them in complementary domains: swimming and athletics. For swimming, we show how robust decision rules on pose statistics can detect different motion events during swim starts, with a $F_1$ score of over $91\%$ despite limited data. For athletics, we use a convolutional sequence model to infer stride-related events in long and triple jump recordings, leading to highly accurate detections with $96\%$ in $F_1$ score at only $\pm 5$ms temporal deviation. Our approach is not limited to these domains and shows the flexibility of pose-based motion event detection.
\end{abstract}

%%%%%%%%% BODY TEXT
\section{Introduction}
Performance analysis of athletes in sports is increasingly driven by quantitative methods, with recording, monitoring and analyzing key performance parameters on a large scale.
This is facilitated by the availability of video and other sensory hardware as well as the opportunities for automation with recent advances in machine learning.
Video based method are of special interest, where athletes are tracked by one or multiple cameras to infer parameters and statistics in individual and team sports. This form of external monitoring does not rely on direct sensor instrumentation of athletes \cite{Fasel18, Ismail18}, which in turn could affect the athletes' performance or limit measurements to very specific training sites.

In this work we propose a vision-based system specifically for event detection in athlete motion.
The main objective is to detect points in time, where characteristic aspects of an athlete's motion occur.
Our focus lies on practical solutions that can be adopted to different sports, event definitions and visual environments.
In recent years, many approaches to visual (motion) event detection, or more general to action recognition, are full-stack vision models \cite{Giancola18, Luvizon18, Victor17}.
They directly try to solve the mapping from images or video clips to very specific motion or action related objectives.
The drawback of this end-to-end paradigm is that transferring and adapting to other domains usually requires a large amount of annotated image or video material and carefully tuning many parameters.
%This is impractical, when adaption to many or very specific domains is required.
In contrast, we encourage the effective decoupling of the video-to-motion-event translation, by using temporal human pose sequences as a suitable intermediate motion representation.
This allows us to built upon existing state-of-the-art vision models, in this case for human pose estimation, as much as possible.
The compact pose sequences serve as the basis for much smaller and simpler event detection models.
%The scope of our work is limited to individual sports, where
We address the problem of detecting motion events for performance analytics in two domains: swimming as well as long and triple jump. In both cases the goal is to precisely extract timestamps of certain motion aspects that in turn can be used to infer timing and frequency parameters.
We purposely choose rather different domains to show the generality of our approach.

\begin{figure*}[t]
\begin{center}
   \includegraphics[width=0.95\linewidth]{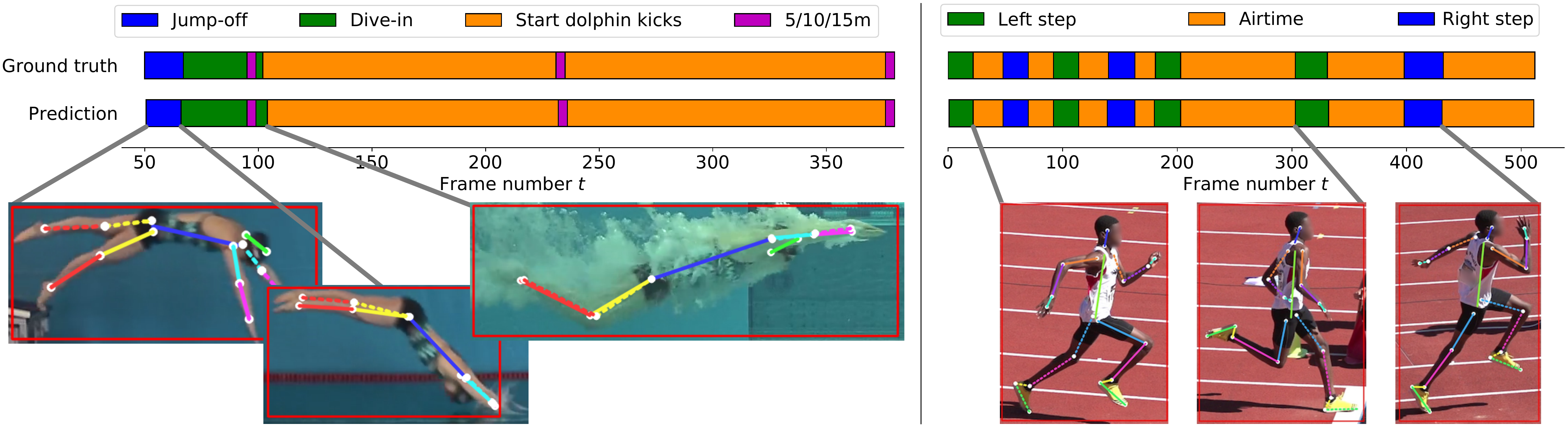}
   \caption{Examples of event detections obtained with our method.
     %The discrete events segment the overall motion into distinct phases.
     Left: Events detected in swim start recordings, including the jump-off, the dive-in and the start of dolphin kicks.
     Right: Detections of begin and end of ground contact for a triple jump athlete.}
\end{center}
\label{fig:qualitative}
\end{figure*}

For swimming, we consider an instrumented swimming pool with synchronized cameras mounted above and under water at the side of the pool.
The four cameras cover the first 20 meters of the pool and are specifically used for swim start training.
They record the jump from the starting block, the dive-in into the water and the initial under water phase of a single athlete under test.
Due to the fixed viewpoints we have strong prior knowledge of what motion will occur where, \ie in which camera, and in which order.
At the same time, only a limited number of recordings with labeled temporal events is available.
For this scenario, temporal motion events can be detected from 2D pose sequences by directly extracting \emph{pose statistics} and deciding on event occurrences with simple \emph{heuristics}.
We show that despite the lack of data, domain specific knowledge can be employed to still obtain robust event detections. 

For long- and triple jump, we use recordings of individual athletes with a single pannable camera.
The camera is positioned at the side of the running track and records the run-up, the jump(s) and the landing in the sand pit.
%Since we mainly focus on the run-up and the jump phase, the recordings are exemplary for many athletics sports.
Compared to swimming, this scenario is rather complementary.
The recordings cover many different tracks during training and competitions, leading to varying viewpoints and scales.
Due to the pannable camera, the viewpoint is also changing during a video. Additionally, there is more variability in timing and location of the observed motion, depending on the camera tracking, the length of the run-up and the step frequency. We show how a moderately sized set of event-annotated recordings can be used to learn a \emph{CNN-based sequence-to-sequence translation}. The idea is to map estimated 2d pose sequences extracted from the recordings to a \emph{timing estimate of event occurrences}. We enhance this approach with a novel pose normalization and augmentation strategy for more robust and stable estimates. Figure~\ref{fig:qualitative} shows examples of video recordings and detected events.

%------------------------------------------------------------------------
\section{Related work}
A lot of prior literature focuses on the tasks of visual motion event detection and action segmentation in sports. We briefly review existing approaches based on their methodology, and in particular the type of motion representation they use, if any.
One possibility is to use semantically rich motion capture data to infer action type and duration of an athlete.
\cite{Vicente16} use 3D pose data from a RGB-D sensor to identify characteristic poses in martial arts with a probabilistic model.
Similarly, \cite{DeDios13} use motion capture data to segment the motion during general physical exercises into different actions.
Limited to regular monocular video recordings, \cite{Li10} describe the usage of low-level video segmentation features in high-diving. They fit a simple body model on the segmented video areas and use a probabilistic model to infer the most likely action.
\cite{Wu02} use estimated athlete velocities from motion segmentation for a high-level semantic classification of long jump actions.
More similar to our approach, \cite{Yagi18} use noisy 2D pose estimates of athletes in sprints. They align lane markers with the 2D poses to infer step frequencies, but with a much lower precision compared to our work.
\cite{Lienhart18} use pose similarity and temporal structure in pose sequences to segment athlete motion into sequential phases, but again with limited temporal precision.
Directly related to our work, \cite{Einfalt19} introduces a deep learning based framework for sparse event detection in 2D pose sequences. Parts of our work build and improve on that foundation.
Lastly, there are multiple approaches that omit an intermediate motion representation and directly infer events from video recordings. \cite{Giancola18} use temporally pooled CNN features directly from soccer broadcasts to classify player actions.
\cite{Sha14} use highly specific body part detectors for swimmers in hand-held camera recordings.
They extract video frames with certain body configurations that mark the start of a swimming stroke.
\cite{Hakozaki2018, Victor17, woinoski20} consider a similar task, but propose a video-based CNN architecture to detect swimming athletes in specific body configurations.
Our work shares the notion of event detection based on body configuration, but we aim at a more modular and flexible solution due to the intermediate pose representation.

\section{Method}
The main motivation behind our approach is to use the human pose as a compact description of an athlete's motion over time.
It decouples the highly specific objective of motion event detection in particular sports from the low-level problem of inferring information from video data directly.
Given a video of length $N$, our goal is to describe it with a pose sequence of the athlete of interest.
Each pose $p \in \mathbb{R}^{K \times 2}$ is the configuration of $K$ 2D keypoints in a specific video frame that describe characteristic points of the human body.
Depending on the body model, the $K$ keypoints usually coincide with joints of the human skeleton.
Each keypoint is represented by its image coordinates.
From a suitable camera viewpoint, such a sequence of 2D keypoints configurations captures the essence of motion of the human body.
In the following, we describe our proposed approach to track a single athlete and his pose over time and to map the resulting pose sequence to the desired motion events.

\subsection{Motion representation with pose estimates}
\label{sec:motion_representation}
In order to infer the pose of the athlete of interest in every video frame, we build upon the vast and successful work on human pose estimation in images.
The CNN architectures for human pose estimation that emerged over the last years have reached levels of accuracy that enable their direct usage in practical applications, including sports.
In this work we use a modified variant of Mask R-CNN \cite{He17}, fine-tuned on sampled and annotated video frames from our application domains.
%We acknowledge the fact that the performance of Mask R-CNN in human pose estimation is surpassed by the most recent architectures (REF).
The main advantage of Mask R-CNN lies in the single, end-to-end pipeline for multi-person detection and per-person pose estimation.
% The vast majority of other pose estimation pipelines handles this top-down approach with separate CNN architectures (\eg (REF)).
From a practical point of view it is easier to implement, train and embed only a single CNN into an application.
We use the common version of Mask R-CNN with a ResNet-101 \cite{He16} and Feature Pyramid Network (FPN) resolution aggregation \cite{Lin17}.
We additionally evaluate a high resolution variant of Mask R-CNN \cite{Einfalt19}, which estimates keypoints at double the usual resolution.
%At the same time, Mask R-CNN has the run-time advantage of sharing a large part of CNN feature calculation between person detection and human pose estimation.
%One distinctive disadvantage of the original Mask R-CNN architecture is its low spatial output resolution.
%After detecting potential regions of interest, Mask R-CNN subsamples spatial ResNet features for each region into a fixed rectangular grid of $14 \times 14$. Only at the very end the resolution is doubled by a learned up-sampling via deconvolution. The final detection maps for every keypoint have an effective resolution of $28 \times 28$ with respect to the rectangular image region containing the person.
%It is at most half of the resolution that comparable architectures for 2D human pose estimation employ (REF) and impairs keypoint detection with high spatial precision.
%In order to counter this drawback we adopt a modified version of Mask R-CNN, where an additional CNN branch refines the initial low resolution output (REF). It operates on twice the spatial resolution and leads to an effective output resolution of $56 \times 56$. Our experiments confirm that this leads to a distinct improvement in spatial precision of keypoint detections and in turn to improved event detections.

\subsection{Tracking and merging pose sequences}
Recordings from individual sports typically depict multiple persons: the athlete of interest, as well as other athletes and bystanders. 
Our swimming recordings often show additional swimmers in the background. Videos from athletics show observers and other athletes surrounding the running track. We therefore need to track the athlete of interest by assigning the correct detection and pose to each video frame.
% TODO: Maybe remove ?!
% Pose tracking over time is still an active research field. Despite many efforts, truly end-to-end pose tracking systems are still too resource and data intensive to outperform per-frame pose estimation and subsequent tracking.
Compared to general pose tracking, \ie finding pose sequences for all people in a video, we only need to find a pose sequence for a single athlete. We therefore propose a generic and adaptable tracking strategy that can include domain knowledge about the expected pose and motion of the athlete, with a simple \emph{track, merge and rank} mechanism.

\subsubsection{Initial pose tracks}
We start by processing a video frame-by-frame with Mask R-CNN.
During fine-tuning to our application domains, we train Mask R-CNN to actively suppress the non-relevant persons.
However, this is not always possible, as the athlete of interest can be similar to other persons in a video frame with respect to appearance and scale. Therefore, we obtain up to $D$ person detections in every frame.
We denote the detection candidates at frame $t$ as $\mathbf{c}_t = \left \lbrace d_{t,1}, \dotsc, d_{t,D} \right \rbrace$.
Each detection is described by $d_{t,i} = \left ( x_{t,i}, y_{t,i}, w_{t,i}, h_{t,i}, s_{t,i} \right )$, with the center, width, height and detection score of the bounding box enclosing the detected person.
Each detected person has its corresponding pose estimate $p_{t,i}$.
Given $\mathbf{c}_t$ for every video frame, we want to find the detection (and pose) track belonging to the athlete of interest.

We start to build initial tracks by linking temporally adjacent detections that are highly likely to belong to the same person.
We employ an intersection over union (IoU) criterion, since we expect the changes of the athlete's position and scale from one frame to another to be small.
By greedily and iteratively linking detections throughout the video, we gain an initial set of detection tracks $\mathbf{T}_1, \dotsc, \mathbf{T}_L$.
Each track has start and end times $t1$ and $t2$ and consists of sequential detections, with $\mathbf{T}_j = \left ( d_{t1}, d_{t1 + 1}, \dotsc, d_{t2}  \right )$, where
\begin{equation}
d_t \in \mathbf{c}_t \; \forall d_t \in \mathbf{T}_j
\end{equation}
and
\begin{equation}
\iou \left ( d_t, d_{t+1} \right ) > \tau_{\text{IoU}} \; \forall (d_t, d_{t+1}) \in \mathbf{T}_j.
\end{equation}
We use a very strict IoU threshold of $\tau_{\text{IoU}} = 0.75$, since we do not use backtracking to re-link individual detection later.

\subsubsection {Track merging}
Due to imperfect or missing detections we need to merge tracks that are divided by small detection gaps.
Two tracks $\mathbf{T}_i, \mathbf{T}_j$ can only be considered for merging, if they are temporally close, but disjoint, \ie $t_{1,j} - t_{2,i} \in [1, \tau_{\text{gap}}]$.
We apply up to three criteria whether to merge the tracks.
First, there has to be some spatial overlap between the detections at the end of $\mathbf{T}_i$ and the beginning of $\mathbf{T}_j$, \ie an IoU $>0$.
Second, both tracks should contain detections of similar scale.
Since the athlete moves approximately parallel to the image plane, his size should be roughly the same throughout a video, and more so in the two separate detection tracks:
\begin{equation}
\frac{\lvert \mathbf{T}_j \rvert}{\lvert \mathbf{T}_i \rvert} \cdot \frac{\sum_{d \in \mathbf{T}_i} d_w \cdot d_h}{\sum_{d \in \mathbf{T}_j} d_w \cdot d_h} \in [\frac{1}{\tau_\text{scale}}, \tau_\text{scale}]
\end{equation}
Lastly, in the case of swimming, we expect the athlete to always move in the same horizontal direction through the fixed camera views. Both tracks should therefore move in the same horizontal direction:
\begin{equation}
\sgn \left ( d_{t2, i, x} - d_{t1, i, x} \right ) = \sgn \left ( d_{t2, j, x} - d_{t1, j, x} \right )
\end{equation}
After greedily merging the initial tracks according to those criteria, we get our final track candidates $\mathbf{T}_1^\prime, \dotsc, \mathbf{T}_M^\prime$.

\subsubsection {Track ranking}
\begin{figure}[t]
\begin{center}
   \includegraphics[width=0.95\linewidth]{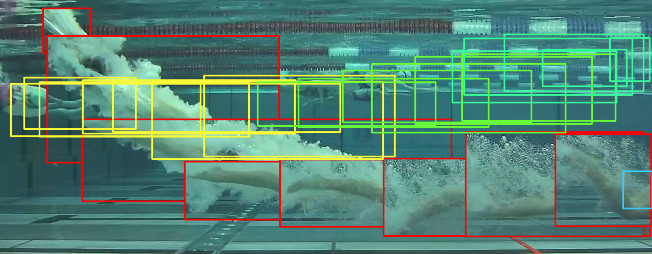}
\end{center}
   \caption{Swimmer detections after track merging in an under water camera. Track ranking selects the correct track (red). Suppressed tracks of background swimmers are superimposed in different colors (yellow $\rightarrow$ blue). Only every 15th detection is shown.
}
\label{fig:tracking}
\end{figure}

In order to select the detection track belonging to the athlete of interest, we rank all final track candidates according to the expected pose and motion of the athlete.
We impose four different rankings $r_k (\mathbf{T}_i^\prime) \rightarrow [1, M]$ on the tracks: (1) The largest bounding box, (2) the highest average detection score, (3) the longest track (long and triple jump only) and (4) the most horizontal movement (swimming only).
%\begin{enumerate}
%\item Largest detection size (\ie average bounding box size)
%\item Highest average detection score
%\item Longest track (long and triple jump only)
%\item Largest average horizontal frame-to-frame movement (swimming only)
%\end{enumerate}
The final track is selected according to the best average ranking:
\begin{equation}
\mathbf{T}_{\text{final}} = \arg \min_{\mathbf{T}_i^{\prime}} \sum_{k=1}^{4} r_k \left ( \mathbf{T}_i^\prime \right )
\end{equation}
It also determines the pose sequence of the athlete. Figure~\ref{fig:tracking} depicts an example for track ranking in an under water swimming recording.

\subsection{Event detection on pose statistics}
The pose sequence allows us to extract timestamps that mark important motion events.
For swimming, we propose to directly identify events based on pose statistics, as the number of training examples is limited and hinders learning a pose-to-event model purely from data. We employ robust decision rules on an observed pose sequence that leverage the fixed temporal structure of the expected motion and the known and fixed camera setting.
In our case, we detect three different categories of events. (1) Position-based events occur when the athlete reaches a certain absolute position in a camera view. In our case, we detect the timestamps of the athlete reaching fixed horizontal distance markings in the calibrated under water cameras. They are used to measure the time for the athlete to cover the first five, ten and fifteen meters. The detection is simply based on the estimated head position surpassing the distance markings. (2) Presence-based events occur when a specific body part is visible for the first or last time in a camera view. Specifically, we detect the begin of the dive-in after jumping from the starting block. It is defined as the first timestamp where the athlete's head touches the water in the  above water camera view. We identify it by a clear reduction in confidence of the head detection due to its invisibility. (3) Pose-based events are defined by the athlete or a subset of his body parts appearing in a certain pose configuration. In our specific scenario we detect the timestamp of last contact of the foot and the starting block as well as the first under water dolphin kick after dive-in.
The former can typically be inferred from the knee angle being maximal when the foot leaves the starting block
% And while the former is essentially a position based event with respect to the foot and the starting block, the single above water camera is subject to small viewpoint changes and the position of the starting block is unkown. We therefore use the auxiliary characterstic of the knee angle being maximal when the foot leaves the starting block.
The dolphin kick can be inversely detected by the smallest knee angle after dive-in.

For robust event detections we require the above-mentioned pose characteristics (\eg low detection confidence or small knee angle) to be present for multiple frames to avoid missdetections due to single, erroneous pose estimates. Additionally, we enforce all event detection to be in the correct order and to appear in the correct camera view.
%Figure~\ref{fig:qualitative} shows examples for event detections.

\subsection{Event detection via sequence-to-sequence translation}
In the domain of long- and triple jump recordings, our goal is to precisely detect stride related events.
Specifically, we want to detect every begin and end of ground contact of an athlete's foot on the running track.
Given a video of length $N$, we denote the set of event occurrences of type $c \in C$ as $\mathbf{e}_c = (e_{c,1}, \dotsc, e_{c,E})$.
Each occurrence $e_{c,i}$ is simply a video frame index. We do not explicitly distinguish between ground contact of the left and right foot, \ie $C = \lbrace \textit{step begin}, \textit{step end} \rbrace$.
%Figure~\ref{fig:qualitative} shows examples for these events.
In contrast to swimming, directly inferring these events from 2D pose sequences with simple decision rules is difficult due to varying camera viewpoints and the a priori unknown number of event occurrences \cite{Yagi18}. Instead, we use a set of annotated videos and the extracted pose sequences to train a CNN for event inference.

Based on prior work \cite{Dauphin17, Gehring17, Li18}, we adopt the notion of a temporal convolutional neural network that translates compact input sequences into a target objective. We build upon the concept of representing discrete event detection in human motion as a continuous translation task \cite{Einfalt19}. Given the input pose sequence $\mathbf{p}$, the objective is to predict a timing indicator $f_c(t)$ for every frame index $t$, that represents the duration from $t$ to the next event occurrence of type $c$:
\begin{equation}
  f_c (t)  = \min_{\substack{e_{c,i} \in \mathbf{e}_c \\ e_{c,i} \geq t}}  \frac{e_{c,i} - t}{t_{\max}}.
\end{equation}
The duration is normalized by a constant $t_{\max}$ to ensure that the target objective is in $[0, 1]$. Analogously, a second objective function $b_c(t) \in [-1, 0]$ is defined to represent the backwards (negative) duration to the closest previous event of type $c$. Every event occurrence is identified by $f_c(t) = b_c(t) = 0$. This type of objective encoding circumvents the huge imbalance of event- and non-event examples. It provides a uniformly distributed label with semantical meaning to every frame index and every input pose.

\subsubsection{Network architecture}

\begin{figure}[t]
\begin{center}
   \includegraphics[width=0.92\linewidth]{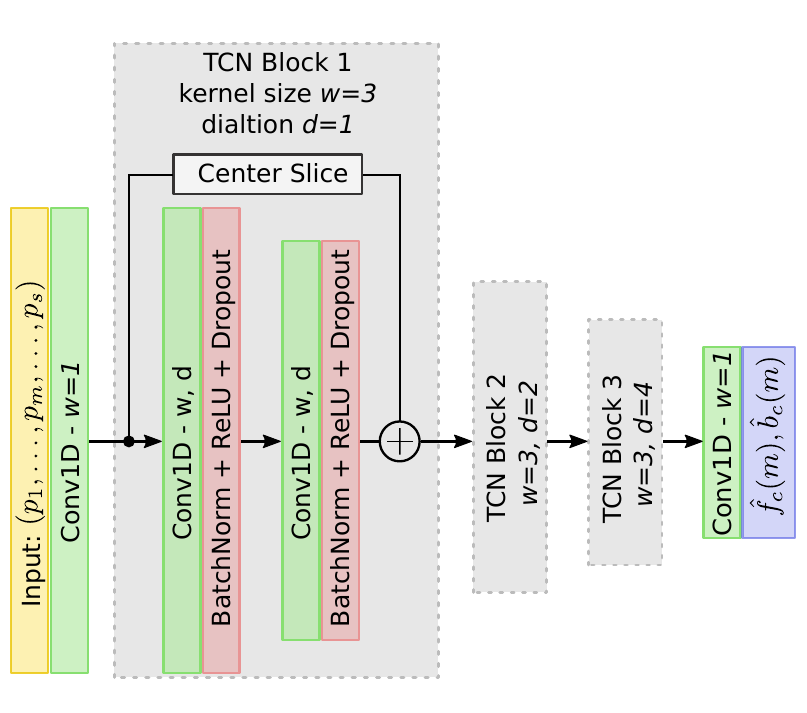}
\end{center}
   \caption{Overview of our CNN architecture for translating pose sequences to event timing indicators. All convolutions except the last have $n=180$ kernels and do not use padding. The residual connection in each TCN block slices its input along the temporal axis to obtain matching dimensions.}
\label{fig:tcn}
\end{figure}

The proposed CNN architecture for learning this sequence translation task follows the generic temporal convolutional network architecture (TCN) \cite{Bai18}. It is designed to map a compact sequential input via repeated convolutions along the temporal axis to a target sequence. The network consists of $B$ sequential residual blocks. Each block consists of two convolutions, each followed by batch normalization, rectified linear activation and dropout. It is surrounded by a residual connection from block input to block output. Each block uses dilated convolutions, starting with a dilation rate of $d=1$ and increasing it with every block by a factor of two.
%This enables a large temporal receptive field while keeping a manageable network depth and parameter number.
The output of the final block is mapped to the required output dimension with a non-temporal convolution of kernel size one. In our case, we train a single network to jointly predict the forward and backward timing indicators for both event types.
The network does not require the poses from an entire video to infer the frequently repeating stride related events. We limit the temporal receptive field of the network with $B=3$ blocks and convolution kernels of size $w=3$ along the temporal axis. This leads to a temporal receptive field of $s=29$ time steps. Additionally, the TCN blocks only use valid convolutions without any zero-padding \cite{Pavllo19}. Figure~\ref{fig:tcn} gives an overview of the architecture.

\begin{figure}[t]
\begin{center}
   \includegraphics[width=0.95\linewidth]{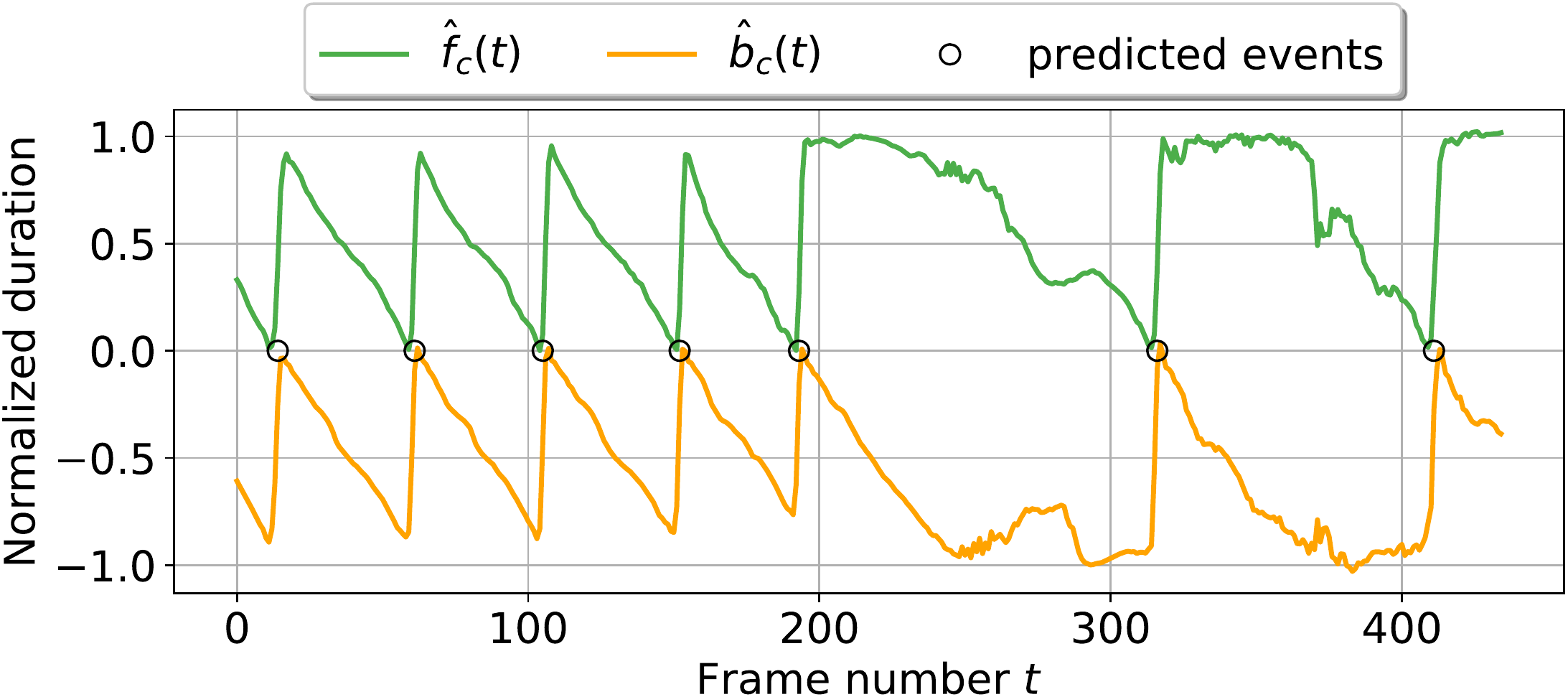}
\end{center}
   \caption{Raw output of our CNN-based pose sequence translation for the last seven \textit{step begin} events in a triple jump video. Events are extracted from the predicted timing indicators, with $\hat{f}_c(t) \approx \hat{b}_c(m) \approx 0$. The example matches the one in Figure~\ref{fig:qualitative}.}
\label{fig:sequence_translation}
\end{figure}

During training, we randomly crop pose sequences of size $s$ from the training videos. The output for these examples consists of only the predictions at the central sequence index $m=\lceil s/2 \rceil$. Additionally, sequences in a minibatch are sampled from different videos to avoid correlated batch statistics \cite{Pavllo19}. We train the network using a smooth-$L_1$ (or Huber) loss \cite{Girshick15, Huber73} on the difference between predicted timing indicators $\hat{f}_c(m), \hat{b}_c(m)$ and the ground truth. Figure~\ref{fig:sequence_translation} depicts the exemplary output on a triple jump video.

\subsubsection{Pose normalization}
\label{sec:pose_rep}
Due to its  fully convolutional nature, the input to our network can be a sequence of pose estimates of variable length. A pose at time $t$ is represented as a 1D vector $\in \mathbb{R}^{3K}$. In contrast to \cite{Einfalt19} we include the detection score of each keypoint, as it contains information about detection uncertainty and possible keypoint occlusion. We mask keypoints with zeros if their score is below a minimal value $c_{\min}$.

The input poses are normalized in order to train the network on pose sequences with varying scales and different video resolutions. We analyze three different normalization strategies. Given a video and its pose sequence $\mathbf{p} = (p_1, \dotsc, p_N)$, we denote the normalized pose at time $t$ as $p_t^\prime$. Our first variant normalizes input poses on a global video level. We set
\begin{equation}
p^{\prime}_{t} = \norm \left ( p_t, \mathbf{p} \right ),  \label{eq:global_norm}
\end{equation}
where 
$norm ( p_t, \mathbf{p})$ min-max normalizes the image coordinates in $p_t$ to $[-1, 1]$ with respect to the observed coordinates in  $\mathbf{p}$. This retains absolute motion of athlete and camera throughout the video, but is susceptible to single outlier keypoints. The second variant limits normalization to a minimal temporal surrounding equal to the receptive field of the network, with
\begin{equation}
p_t^{\prime} = \norm \left ( p_t, \left ( p_{t-m}, \dotsc, p_{t+m} \right ) \right ). \label{eq:local_norm}
\end{equation}
Due to its locality it is more robust and largely removes absolute motion on video level. However, since each pose is normalized with respect to its own, different temporal surrounding, adjacent normalized poses $p_t^\prime, p_{t+1}^\prime$ are no longer directly comparable. Finally, the third variant tries to combine the advantages of the former strategies with a sequence based normalization. We manually extract overlapping sub-sequences of size $s$ from the video and normalize all poses within equally. This effectively changes the operation of our network during inference. In order to compute the network output at a single time index $t$, we extract the surrounding sub-sequence and jointly normalize it:
\begin{equation}
  p_i^{\prime} = \norm \left ( p_i, \left ( p_{t-m}, \dotsc, p_{t+m} \right ) \right ) \;  \forall_{i=t-m:t+m} \label{eq:seq_norm}
\end{equation}
This retains a local normalization and keeps adjacent poses inside of a sub-sequence comparable. The drawback is that for each output, a differently normalized sub-sequences has to be processed. It removes the computational efficiency of a convolutional network.
However, the small size of the network still keeps processing time for an entire video pose sequence in the order of seconds.

\subsubsection{Pose sequence augmentation}
During training and inference, the input poses to our sequence translation network are estimates themselves, including missing, wrong or imprecise pose estimates. The network needs to learn to cope with imperfect data.
% while the number of training videos is limited, however.
We therefore want to enhance the training data to reflect the variability and the error modes of pose estimates that the model encounters during inference.
Simply augmenting the training data by randomly perturbing input poses is not convincing.
Adding pose and motion agnostics noise might introduce error modes that are not present during test time.
Our proposal is to extract pose sequences from the training videos multiple times with slightly different pose estimation models. In our case we use original Mask R-CNN as well as its high spatial precision variant.
Additionally, we use multiple different checkpoints from the fine-tuning of both models, each leading to unique pose sequences with potentially different, but realistic error modes.
%This form of pose sequence augmentation only requires the training of additional pose estimation models and the repeated extraction of pose sequences.
%It comes at no additional need of labeled data.

%------------------------------------------------------------------------
\section{Experimental setting}
We evaluate our approach to human pose estimation, athlete tracking and subsequent event detection on real world video recordings. For swimming, our dataset consists of 105 recordings of swim starts, each comprising of four synchronized camera at $50$ fps: one above water and three under water. All recordings are annotated with the event types $C=$ \{\textit{jump off}, \textit{dive-in}, \textit{first kick}, \textit{5m}, \textit{10m}, \textit{15m}\}. Each event occurs exactly once per recording. We use 23 recordings for optimization of tracking and event detection parameters.
For long and triple jump, we use 167 monocular recordings at 200 fps from various training and competition sites, of which 117 are used for training and validation. They are labeled with event occurrences of $C=\lbrace \textit{step begin}, \textit{step end}\rbrace$. Due to the repetitive motion, each event type occurs nine times per video on average.

%\subsection{Extraction of 2D pose candidates}
\textbf{Extraction of 2D pose candidates}$\;$We use Mask R-CNN, pre-trained on COCO \cite{Coco14}, and separately fine-tune the model on sampled and annotated video frames from both domains. For swimming, we use 2500 frames, annotated with a standard $K=14$ body model. For long and triple jump, a total of 3500 frames are annotated with $K=20$ keypoints, specifically including the feet of the athlete. The Mask R-CNN model is fine-tuned with a batch size of $8$ for $140$ epochs, a base learning rate of $0.1$ and a reduction by $0.1$ after $120$ epochs.  We process all videos and camera views frame-by-frame and extract the $D=3$ highest scoring athlete detections and their pose estimates.

%\subsection{Athlete tracking}
\textbf{Athlete tracking}$\;$
Given the multiple detections per video frame, we apply our athlete tracking strategy to obtain a single detection and pose sequence per video. For swimming, each camera view is processed independently and therefore has its own pose sequence. We speed up pose inference and tracking by only processing a camera view if the athlete already appeared in the previous camera.
All tracking parameters $\tau$ are optimized with a grid search on the training videos. For long- and triple jump videos, tracking is optimized for athlete detection performance on the pose-annotated video frames. For swimming, we jointly optimize tracking parameters and the hand-crafted event detection decision rules directly for event detection performance.

%\subsection{Long and triple jump event timing estimation}
\textbf{Long and triple jump event timing estimation}$\;$
The temporal convolutional network for event timing prediction in long and triple jump is trained on the inferred 2D pose sequences from all training videos They contain $2167$ step events, equally distributed among \textit{step begin} and \textit{step end}. We extract training sequences of length $s=29$ from the per-video pose sequences, leading to a total training set of 65k different input pose sequences. The network is trained with a batch size of $512$, dropout rate of $0.1$ and a base learning rate of 1e-2 using the Adam optimizer for 20 epochs or until convergence. The learning rate is reduced by $0.3$ after $10$ epochs. Discrete event occurrences are extracted as shown in Figure~\ref{fig:sequence_translation}.

%\subsection{Evaluation protocols}
\textbf{Evaluation protocols}$\;$
After extracting event predictions for each recording in the test set, we exclusively assign every prediction to its closest ground truth event. A prediction is correct if its absolute temporal distance to the assigned ground truth does not exceed a maximum frame distance $\Delta t$. We report detection performance at maximum frame distances of $\Delta t \in [1,3]$. We do not consider $\Delta t = 0$, since even annotations by humans often deviate by one frame. At the same time, event detection performance usually saturates at $\Delta t = 3$ despite the different frame rates in swimming and long and triple jump. Given a maximum frame distance, we report \textit{precision}, \textit{recall} and the combined $F_1$ score. Figure~\ref{fig:qualitative} shows qualitative examples of detected events.

We additionally measure pose estimation performance with the standard \textit{percentage of correct keypoints} (PCK) metric \cite{Sapp13} on a pose-annotated test set of 600 frames for swimming and 1000 frames for long and triple jump. For athlete detection, we report the standard \textit{average precision} (AP) metric \cite{Coco14} at a required bounding box IoU of $0.75$ on the same set of frames. For swimming, we also report the \textit{false positive rate} (FPR) of our tracking approach on a separate set of video frames that do not depict the athlete of interest. Note that all metrics are reported as a percentage.

% ------------------------------------------------------------------------
\section{Results}
\subsection{Per-frame pose estimation}

Table~\ref{tab:pose_estimation} shows the pose estimation results on test set frames.
We report these results as reference for the pose estimation fidelity on which the subsequent tracking and event detection pipeline operates.
The table shows PCK results at thresholds of $0.05$, $0.1$ and $0.2$, which correspond to a very high, high and low spatial precision in keypoint estimates. With the original Mask R-CNN architecture we achieve PCK values of $70.0$,  $88.8$ and $95.1$ for long and triple jump. Especially the value at PCK@$0.1$ indicates that the model produces reliable and precise keypoint estimates for the vast majority of test set keypoints. The high resolution variant of Mask R-CNN leads to another gain in PCK of up to $+2.4$ at the highest precision level.
%The improvement only comes from an architectural modification, but otherwise no change in training data.
For swimming, we achieve a base result of $50.7$, $78.0$ and $92.5$ for the respective PCK levels, with a notable drop in high precision keypoint estimation compared to the athletics videos. The main difference is the aquatic environment, the static cameras leading to truncated poses and the lower number of annotated training frames. Especially the underwater recordings are known to pose unique challenges like the visual clutter due to bubbles and low contrast \cite{zecha19} (see Figure~\ref{fig:tracking}). High resolution Mask R-CNN leads to a small gain in high precision PCK of up to $+1.1$, but otherwise seems to suffer from the same difficulties.
% as standard Mask R-CNN does in this environment.
\begin{table}[]
\begin{center}
\begin{tabular}{@{}lcc@{}}
\toprule
                                 & Swimming                                & \multicolumn{1}{l}{Long/triple jump} \\ \midrule
                                 & \multicolumn{2}{c}{PCK@0.05 / 0.1 / 0.2}                                       \\ \midrule
\multicolumn{1}{l|}{Mask R-CNN}  & 50.7 / 78.0 / \textbf{92.5} & 70.0 / 88.8 / \textbf{95.1}                   \\
\multicolumn{1}{l|}{+ high res.} & \textbf{51.8} /\textbf{ 78.4} / 92.4 & \textbf{72.4} / \textbf{89.1} / \textbf{95.1}                   \\ \bottomrule
\end{tabular}
\end{center}
\caption{Results on per-frame human pose estimation on test videos. We compare the original Mask R-CNN architecture and a high resolution variant.}
\label{tab:pose_estimation}
\end{table}

\subsection{Athlete tracking}

Table~\ref{tab:tracking} shows results on athlete bounding box detection.
The reported AP is measured on the same set of test set frames that is used in pose estimation evaluation.
These frames all are positive examples, \ie the athlete of interest is known to be seen.
We evaluate our tracking approach by processing the complete video recordings and filtering for the detections in those specific frames.
We compare this result to an optimistic baseline, where we directly apply Mask R-CNN to only those frames, avoiding the necessity for tracking.
For long and triple jump, tracking is on par with the optimistic baseline. Despite tracking being guaranteed to find a single, temporally consistent detection sequence, it is still able to retain the same recall. But with an average precision of $97.9$, the detection quality of Mask R-CNN alone is already very high for this domain. In contrast, the detection performance on swimming is considerably lower. Tracking slightly surpasses the baseline by $+0.7$ with an AP of $76.0$. It retains the recall of the optimistic baseline and also improves the suppression of irrelevant detections in the positive test set frames.
One main difference to the long and triple jump recordings is the multi-camera setup, where the athlete is usually only visible in one or two cameras at the same time.
Tracking therefore also needs to suppress detections in a large number of negative frames, where the athlete of interest is not visible.
Table~\ref{tab:tracking} shows the false positive rate (FPR) on our set of negative frames. With tracking we obtain false detections in $2.1$ percent of the negative frames, which is considerably lower than the $6.2$ FPR of the baseline.

\begin{table}[]
\begin{center}
\begin{tabular}{@{}lccc@{}}
\toprule
                                & \multicolumn{2}{c}{Swimming}           & \multicolumn{1}{l}{Long/triple jump} \\ \midrule
                                & AP$_{0.75}$ & FPR & AP$_{0.75}$                          \\ \midrule
\multicolumn{1}{l|}{Mask R-CNN} & 75.3        & 6.2 & \textbf{97.9}                                 \\
\multicolumn{1}{l|}{+ tracking} & \textbf{76.0}        & \textbf{2.1} & \textbf{97.9}                                 \\ \bottomrule
\end{tabular}
\end{center}
\caption{Results on athlete detection in positive (AP) and negative (FPR) frames with our tracking strategy. Performance is compared to an optimistic baseline with per-frame Mask R-CNN results.}
\label{tab:tracking}
\end{table}

\subsection{Event detection in swimming}

Table~\ref{tab:simming_events} shows the results on event detection in swimming recordings when applying our hand-crafted decision rules on pose sequences obtained via Mask R-CNN and tracking. We only report the recall of event detections, as exactly one event is detected per type and recording.
%The results on events defined by the underwater distance markings are averaged, as they are nearly identical.
At $\Delta t=1$ we already achieve a recall of at least $91.3$ for the jump-off, dive-in and the distance-based events. The majority of remaining event occurrences is also detected correctly when we allow a frame difference of $\Delta t=3$, with a recall of at least $97.1$. This shows that our approach of using hand-crafted decision rules on pose statistics is capable of precisely detecting the vast majority of those event types.
%This is especially remarkable when recalling the huge difference in precise person and keypoint detection compared to the long and triple jump recordings.
%Using appropriate decision rules that do not require perfect detections for all keypoints in a pose proves to be a valid approach.
The only exception is the recall for the first dolphin kick, which saturates at $89.4$ even for frame differences $\Delta t > 3$. The respective decision rule thus sometimes generates false positives that are distant from the actual event occurrence. We observed that nearly all of those false positives are detections of a small knee angle during the second dolphin kick. The main cause for this seem to be the unstable detections for hip, knee and ankle keypoints during the first kick when large amounts of bubbles are in the water from the dive-in.

\begin{table}[]
\begin{center}
\begin{tabular}{@{}lcc@{}}
\toprule
\multicolumn{1}{r}{Recall at}      & $\Delta t=1$ & \multicolumn{1}{l}{$\Delta t =3$} \\ \midrule
\multicolumn{1}{l|}{Jump-off}     & 91.4                     & 97.1                    \\
\multicolumn{1}{l|}{Head dive-in} & 92.9                     & 98.6                    \\
\multicolumn{1}{l|}{First kick}   & 84.8                     & 89.4                    \\
\multicolumn{1}{l|}{5m/10m/15m}   & 91.3                     & 99.0                    \\ \bottomrule
\end{tabular}
\end{center}
\caption{Results on event detection in swimming recordings at different temporal precision levels $\Delta t$.}
\label{tab:simming_events}
\end{table}

\subsection{Event detection in long and triple jump}

Table~\ref{tab:athletics_events} shows the results on event detection in long and triple jump recordings.
Our base model operates on pose sequences obtained with high resolution Mask R-CNN and our temporal athlete tracking. Input poses are sequence normalized (Equation~\ref{eq:seq_norm}). No additional data augmentation is applied. The results on the event types \textit{step begin} and \textit{step end} are averaged, as they do not show distinct differences. The base model already achieves a $F_1$ score of $95.5$ even at the strictest evaluation level with $\Delta t = 1$. It correctly detects the vast majority of the step-related events, with true positives having a mean deviation of only $2ms$ from the ground truth. The $F_1$ score improves to $98.4$ at the more relaxed evaluation with $\Delta t = 3$. With a precision of $99.5$, the only remaining error modes are false negatives, \ie events that simply do not get detected no matter the temporal precision $\Delta t$.
We also compare our base model to a variant that uses pose estimates from a regular Mask R-CNN model without the high resolution extension. The loss in high precision pose estimation from Table~\ref{tab:pose_estimation} translates to a reduction of $-1.1$ in $F_1$ score at $\Delta t=1$. There are only marginal differences at lower temporal precision. This shows that despite regular Mask R-CNN already achieving very reliable pose estimates in this domain, additional improvements in keypoint precision can still be leveraged by our pose sequence model.

We additionally explore the effects of different pose normalizations, as proposed in Section~\ref{sec:pose_rep}. Table~\ref{tab:athletics_events} (mid) shows a large drop in $F_1$ score of up to $-5.8$ when input poses are normalized globally (Equation~\ref{eq:global_norm}). This clearly indicates that retaining information about absolute motion in a video hinders precise event detection in this domain.
Consequently, performance largely recovers when using local pose normalization from Equation~\ref{eq:local_norm}. But the fact that poses in an input sequence are all normalized differently still leads to loss of up to $-1.5$ in $F_1$ score compared to the base model.
It proves that the sequence-based normalization indeed combines the advantages of the other two normalization methods, leading to a highly suitable pose representation for event detection.

Finally, Table~\ref{tab:athletics_events} (bottom) shows the result when using pose sequence augmentation.
The CNN-based sequence translation is trained with pose sequences extracted with different pose estimation models. We use three different checkpoints during fine-tuning of original and high resolution Mask R-CNN.
%It results in six times the number of training pose sequences compared to the base model.
This leads to our best performing model, with additional gains for all precision levels $\Delta t$ of up to $+1.2$ in $F_1$ score. The further improvement confirms that augmenting pose sequences with realistic error modes from various pose estimation models is a valid strategy.
If needed, it could even be extended to include pose estimates from entirely different models.
% Overall, the results clearly shows that our proposed pipeline of pose estimation, tracking and sequence translation offers highly reliable event detections that can be used in automated performance analysis.

\begin{table}[]
\begin{center}
\begin{tabular}{@{}lccc@{}}
\toprule
\multicolumn{1}{r}{$F_1$ score at}        & $\Delta t = 1$ & \multicolumn{1}{l}{$\Delta t = 2$} & \multicolumn{1}{l}{$\Delta t = 3$} \\ \midrule
\multicolumn{1}{l|}{Base model}             & 95.5           & 98.1                               & 98.4                               \\
\multicolumn{1}{l|}{w/o high res.}          & 94.4           & 98.3                               & 98.6                               \\ \midrule
\multicolumn{1}{l|}{w/ global norm.}        & 88.6           & 94.5                               & 96.6                               \\
\multicolumn{1}{l|}{w/ local norm.}         & 94.0           & 97.7                               & 98.3                               \\ \midrule
\multicolumn{1}{l|}{w/ pose augmentation} & \textbf{96.7}           & \textbf{98.4}                               & \textbf{98.8}                               \\ \bottomrule
\end{tabular}
\end{center}
\caption{Results on event detection in long and triple jump, with different variants of our CNN-based event detection.}
\label{tab:athletics_events}
\end{table}

%------------------------------------------------------------------------
\section{Conclusion}

In this paper, we have presented a practical approach to motion event detection in athlete recordings. 
%Our main motivation was to use 2D human pose sequences as an intermediate motion representation.
It avoids the need to develop a complete end-to-end vision model for this highly domain-dependent task.
Instead, we build on the state-of-the art in human pose estimation to obtain a compact description of an athlete's motion in a video.
Our first contribution is a flexible tracking strategy for a single athlete of interest that suppresses hardly avoidable missdetections of other athletes and bystanders.
We showed how domain knowledge about appearance and motion can be leveraged to obtain consistent detection and pose sequences.
Our second contribution are two different approaches to event detection on pose sequences.
For swimming, we showed how robust decision rules on pose statistics can already achieve convincing results, despite limited data.
With a sufficient set of annotated event occurrences, we additionally showed how a CNN-based sequence translation can be used to learn event inference in the domain of long and triple jump.
We focused on finding appropriate pose normalization and augmentation strategies, leading to a highly reliable model with hardly any error modes.
Both approaches are not strictly limited to the specific domains we applied them to, but rather show the flexibility of human pose sequences as a foundation for motion event detection.

%\section*{Acknowledgments}
\textbf{Acknowledgments}$\;$
This work was funded by the Federal Institute for Sports Science based on a resolution of the German Bundestag.
We would like to thank the Olympic Training Centers Hamburg/Schleswig-Holstein and Hessen for collecting and providing the video data.

{\small
\bibliographystyle{ieee_fullname}
\bibliography{references}
}

\end{document}